\documentclass{article}

\usepackage{microtype}
\usepackage{graphicx}
\usepackage{subfigure}
\usepackage{booktabs} 

\usepackage{color}

\usepackage{sidecap}

\usepackage{hyperref}

\usepackage[accepted]{icml2024}

\usepackage{amsmath}
\usepackage{amssymb}
\usepackage{mathtools}
\usepackage{amsthm}

\usepackage[capitalize,noabbrev]{cleveref}
\usepackage{float}

\theoremstyle{plain}
\newtheorem{theorem}{Theorem}[section]

\theoremstyle{definition}
\newtheorem{definition}[theorem]{Definition}

\theoremstyle{remark}

\usepackage[textsize=tiny]{todonotes}

\usepackage{physics}

\newcommand{\R}{\mathbb{R}}
\newcommand{\N}{\mathbb{N}}
\newcommand{\Q}{\mathbb{Q}}

\icmltitlerunning{GRaM Workshop at ICML 2024}

\begin{document}

\twocolumn[
\icmltitle{GLAudio Listens to the Sound of the Graph}



\icmlsetsymbol{equal}{*}

\begin{icmlauthorlist}

\icmlauthor{Aurelio Sulser}{equal,yyyy}
\icmlauthor{Johann Wenckstern}{equal,yy}
\icmlauthor{Clara Kuempel}{equal,yyyy}


\end{icmlauthorlist}

\icmlaffiliation{yy}{Department of Mathematics, ETH Zürich, Switzerland}

\icmlaffiliation{yyyy}{Department of Computer Science, ETH Zürich, Switzerland}

\icmlcorrespondingauthor{Aurelio Sulser}{asulser@ethz.ch}

\icmlkeywords{Machine Learning, Deep Learning, Graph Neural Networks}

\vskip 0.3in
]



\printAffiliationsAndNotice{\icmlEqualContribution} 

\begin{abstract}
We propose GLAudio: Graph Learning on Audio representation of the node features and the connectivity structure. This novel architecture propagates the node features through the graph network according to the discrete wave equation and then employs a sequence learning architecture to learn the target node function from the audio wave signal. This leads to a new paradigm of learning on graph-structured data, in which information propagation and information processing are separated into two distinct steps. We theoretically characterize the expressivity of our model, introducing the notion of the receptive field of a vertex, and investigate our model's susceptibility to over-smoothing and over-squashing both theoretically as well as experimentally on various graph datasets.
\end{abstract}

\section{Introduction}
With the advent of the popular architectures GCN \cite{kipf2017semisupervised}, GAT \cite{velickovic2017graph}, and GIN \cite{xu2018powerful}, Graph Neural Networks (GNNs) have emerged as a powerful tool for learning on relational data. Despite the theoretical importance of depth for the expressivity of neural networks, most GNNs encountered in applications are relatively shallow. This is related to two fundamental issues impairing the expressivity of deep GNNs: over-squashing \cite{alon2020bottleneck}, \cite{topping2021understanding} and over-smoothing \cite{li2018deeper}, \cite{oono2019graph}. A comprehensive theoretical framework to assess the representational capabilities of GNNs was introduced in \cite{xu2018powerful}. It established that GNNs can possess a representational power at most equivalent to the Weisfeiler-Leman graph isomorphism test when applied to featureless graphs. This revelation prompts an intriguing inquiry: what types of functions are learnable by GNNs when node features are included? A detailed characterization of these functions for most GNNs remains elusive, instead, the two primary limitations to their expressive power have been extensively explored: over-smoothing and over-squashing. The concept of over-squashing was first introduced by \cite{alon2020bottleneck}. They observed that for a prediction task involving long-range interactions between nodes separated by a distance \( k \), a GNN requires an equivalent number of $k$ layers to capture these interactions effectively. As the \( k \)-th neighborhood of a vertex expands exponentially with \( k \), the exploding volume of data must be compressed into a fixed-size vector at each vertex \( v \), leading to over-squashing. In \cite{digiovanni2023does}, it was explored how over-squashing reduces the class of functions learnable by GNNs. In \cite{di2023understanding}, it was noted that standard GNNs are guided by the heat equation, a smoothing process. For such heat-equation guided GNNs, over-smoothing and over-squashing are intrinsically related to the spectral gap of the graph Laplacian, resulting in an inevitable trade-off between these two issues \cite{giraldo2023oversquashing}. In recent years, there has been a significant effort to introduce new architectures addressing these performance impairments, moving beyond heat-equation-guided GNNs. A promising approach are continuous-time GNNs that are often physically inspired \cite{chamberlain2021grand}, \cite{bodnar2022neural}. One such architecture is GraphCON \cite{rusch2022graphcoupled} that models the node features as controlled and damped oscillators, coupled via the adjacency structure of the underlying graph. Inspired by the great success of the Transformer Model in natural language processing, a completely new approach called Graph Transformers was proposed \cite{kreuzer2021rethinking}, \cite{dwivedi2020generalization}, \cite{rampavsek2022recipe}. Unlike GNNs, graph transformers do not propagate the node features over the connectivity structure of the graph instead the node features are augmented by a laplacian encoding of the node's position in the graph. In that sense, the encoding of the connectivity structure is separated from the learning task, mitigating the phenomena over-smoothing and over-squashing.

\textbf{Contribution.} Continuing this idea of separating the encoding of the connectivity structure from the learning task, we propose GLAudio. GLAudio propagates the node features through the graph network according to the discrete wave equation \eqref{eq:Wave} and then uses sequence learning architectures like LSTM or the Transformer model to learn the node function $\mathbf{Y}_v: \R^{|V(G)| \times d_0} \rightarrow \R^{d_1}$ at each vertex $v \in V(G)$ from the wave signal received at $v$. In that sense, GLAudio separates the feature propagation and signal processing into two distinct steps. Unlike the heat equation, the wave equation preserves the Dirichlet energy. Thus, the node features can be propagated over long distances without smoothing out. Moreover, the compression of information takes place during the encoding of node features and graph structure into a wave signal. Consequently, with increasing resolution the signal received at a vertex in GLAudio should be significantly less compressed compared to the fixed-size vector in standard GNNs. These two facts suggest that over-smoothing and over-squashing are mitigated. We are able to characterize the function class learnable by GLAudio and address these two phenomena.


\section{GLAudio}
\subsection{Wave Signal Encodes Features and Graph Structure}
The discrete wave equation on a graph $G$ with Laplacian matrix $\mathbf{L}$ and initial resting configuration $\mathbf{x} \in \R^{|V(G)|\times d_0}$ reads
\begin{align}\label{eq:Wave}
    \left\{\begin{aligned}
\ddot{\mathbf{X}}(t) &= - \mathbf{L} \cdot \mathbf{X}(t)\\
\mathbf{X}(0) &= \mathbf{x}\quad \dot{\mathbf{X}}(0) = 0.
\end{aligned}\right.
\end{align}
It is well known that the unique solution is given by the continuous-time signal $\mathbf{X}(t) = \cos(\mathbf{L}^{1/2}\cdot t) \cdot \mathbf{x}$. The following theorem proves that this signal encodes much of the information about the features and the graph structure.

\begin{theorem}\label{thm:GraphEncoding}
    Given two graphs $G, H$ on the same vertex set with initial features $\mathbf{x}_{G}, \mathbf{x}_{H}$, then we have for the two corresponding wave signals $\mathbf{X}_{G}(t), \mathbf{X}_{H}(t),$ $\forall t > 0$
    \[\mathbf{\mathbf{X}}_{G}|_{[0,t]} = \mathbf{\mathbf{X}}_{H}|_{[0,t]} \iff \forall n \in \N_0: \mathbf{L}_H^n \cdot \mathbf{x}_{H} = \mathbf{L}_G^n \cdot \mathbf{x}_{G}\]
\end{theorem}
Motivated by this fact, we propose to use the signal $\mathbf{X}(t)$ as an encoding of the graph's features and structure. 

\subsection{Model Architecture}
\textbf{Encoder.} To implement the encoder, we solve the differential equation \eqref{eq:Wave} in discrete time using an implicit-explicit ODE scheme \cite{norsett1987solving}. Let $N$ denote the number of discrete time steps and $T$ the stopping time. Let $h = \frac{T}{N}$ be the step size. We denote the approximation of $\textbf{X}(ih)$ by $\textbf{X}^i$ for $i=0,...,N$. The encoder architecture then reads 
\begin{align}\label{encoder}
\left\{\begin{aligned}
    \textbf{X}^{i+1} &= \textbf{X}^i + h \textbf{V}^{i+1}\\
    \textbf{V}^{i+1} &= \textbf{V}^i - h \textbf{L}\cdot \textbf{X}^i\\
    \textbf{X}^0 &= \textbf{x} \quad \textbf{V}^0 = 0.
\end{aligned}\right.
\end{align} 
where $\textbf{V}^i$ is an auxiliary "velocity" variable. We denote the step function provided by the numerical scheme by $ \hat{\textbf{X}}(\textbf{x},t) = \textbf{X}^i \cdot 1_{[0,h]}(t) + \sum_{i = 2}^N \textbf{X}^i \cdot 1_{((i-1)h,ih]}(t)$. It is well known that as $N$ increases $\hat{\textbf{X}}(\textbf{x},t)$ converges in the function space $L^{\infty}([0,T];\R^{|V(G)| \times d_0})$ to the true solution $\textbf{X}(\textbf{x},t)$ uniformly over all possible initial features of some fixed compact set $C \subseteq \R^{|V(G)| \times d_0}$.

\textbf{Decoder.} For the decoder model, we can use any sequence learning architecture. We have tested RNN decoders, in particular, LSTM \cite{gers2000learning} and CoRNN \cite{rusch2020coupled}. But it might also be interesting to investigate how the Transformer model \cite{vaswani2017attention} or State Space Models \cite{smith2022simplified}, \cite{gu2023mamba} perform as a decoder. For sequence learning architectures, universal approximation theorems have been established \cite{schafer2006recurrent}, \cite{lanthaler2023neural}, \cite{yun2019transformers}, \cite{wang2024state}. In the following, we will make use of the universality of RNNs \cite{schafer2006recurrent} to characterize the expressivity of GLAudio. A simple RNN on an input sequence $\textbf{x}_1, \dots, \textbf{x}_n \in \R^{d_0}$ is given by for all $1 \leq i \leq n$
\begin{align}\label{rnn-def}
    \textbf{s}_i &= \sigma(\textbf{W} \cdot \textbf{s}_{i-1} + \textbf{U} \cdot x_{i}) \\
    \textbf{y}_i &= \textbf{V} \cdot \textbf{s}_{i}, \nonumber
\end{align}
where $\textbf{W} \in \R^{s \times s}, \textbf{V} \in \R^{s \times d_1}, \textbf{U} \in \R^{d_0 \times s}$ and $\sigma$ is a non-linearity, $\textbf{s}_i$ are the hidden states and $\textbf{y}_i$ are the outputs.

\begin{theorem}[\cite{schafer2006recurrent}]\label{thm:RNN-universality}
    For any dynamical system $ \textbf{S}_{i+1} = g(\textbf{S}_i,\textbf{X}_{i+1}), \textbf{Y}_{i+1} = h(\textbf{S}_{i+1}),$
    where $g: \R^{s} \times \R^{d_0} \rightarrow \R^{s}$ measurable and $g: \R^{s} \rightarrow \R^{d_1}$ continuous, there exists an arbitrary good approximation of $\textbf{Y}_n$ by some $\textbf{y}_n$ of the form \eqref{rnn-def} uniformly for any input sequence $\textbf{X}_1, \dots, \textbf{X}_n$ of some fixed compact set $C \subseteq \R^{n \times d_0}$.
\end{theorem}

\begin{table*}[t]
    \caption{Node classification test accuracy obtained on network graph datasets. Means are obtained from 10 random initializations over fixed publicly available train/val/test split. Baselines for GCN, GAT and GraphCON are taken from \cite{rusch2022graphcoupled}.}
    \label{tab:heterophilic_homophilic}
    \vskip 0.15in
    \begin{center}
    \begin{small}
    \begin{sc}
\begin{tabular}{c c c c c c}
    \hline
     \textit{Datasets} & GCN & GraphCON-GCN & GAT & GraphCON-GAT  & \textbf{GLAudio-CoRNN}  \\
     \hline 
     \hline 
     Cora & 0.815 &0.819 &0.818 & 0.832 & 0.795 \\ 
     CiteSeer & 0.719 & 0.729 & 0.714& 0.732 & 0.686\\
     PubMed & 0.778 & 0.788 & 0.787 & 0.795 & 0.781\\
     \hline 
     Texas & 0.551 &0.854 &0.522 &0.822 & 0.802 \\
     Wisconsin & 0.518 & 0.878 & 0.494& 0.857 & 0.831\\
     Cornell & 0.605 & 0.843 & 0.619 & 0.832 & 0.764 \\
    \hline
\end{tabular}
\end{sc}
\end{small}
\end{center}
\vskip -0.1in
\end{table*}

For a detailed list of all hyper-parameters and further configuration options, we refer to Appendix \ref{app:ExperimentDetails}. 

\subsection{Expressivity of GLAudio}

In this section, we provide a thorough characterization of the expressive power of GLAudio. This analysis enables us to articulate how GLAudio potentially alleviates the issues of over-squashing and over-smoothing. To ease the notation, we restrict the discussion to the case $d_0 = 1$. The presented results easily generalize to the case of arbitrary $d_0$. Let $\{\phi_i\}_i$ be an eigenbasis of $\mathbf{L}$. We define the receptive field $\mathcal{R}_v$ of a vertex $v$ to be the set $\{ \phi_i \mid \forall i \in [n]: \langle \mathbf{e}_v, \phi_i \rangle \neq 0 \}$. We call a function $f: \R^n \rightarrow \R^m$ \textit{supported on a linear subspace $V \subseteq \R^n$} if for any $\mathbf{u} \perp V$ we have that $\forall \mathbf{x} \in \R^n: f(\mathbf{x} + \mathbf{u}) = f(\mathbf{x})$.

\begin{theorem}\label{thm:Universality}
    Provided all eigenvalues of $\mathbf{L}$ are unique, if $\mathbf{Y}_v$ is supported on $\mathcal{R}_v$, and $\forall \epsilon > 0, \forall C \subset \mathbb{R}^n$ compact there exists an approximation $\mathbf{y}_N$ according to \eqref{rnn-def} on the input sequence $\textbf{X}^1, \dots, \textbf{X}^n$ such that for all initial configurations $\mathbf{x} \in C, \norm{\mathbf{Y}_v(\mathbf{x}) - \mathbf{y}_N(\textbf{X}^1, \dots, \textbf{X}^N)}_2 \leq \epsilon.$
\end{theorem}

\textbf{Over-smoothing} There exist a number of measures in the literature on over-smoothing in deep GNNs, e.g. measures based on the Dirichlet energy \cite{cai2020note}, \cite{zhao2019pairnorm} or on the mean-average distance (MAD) \cite{chen2020measuring}, \cite{zhou2020towards}. The survey paper \cite{rusch2023survey} gives a unified, rigorous, and tractable definition.

\begin{definition}{\cite{rusch2023survey}}
    Given a sequence of GNNs on a graph $G$, where the $(N+1)$-th GNN differs from the $N$-th GNN by exactly one additional layer. We say that the sequence over-smooths w.r.t. some node similarity $\mu$ if $\mu(\mathbf{y}^N) < c_1 e^{-c_2N}$, where $\mathbf{y}^N$ is the output of the $N$-th GNN and $c_1, c_2>0$.
\end{definition}

Note that in our model, as we increase $N$ for fixed stopping time $T$, the input signal $\hat{\textbf{X}}(\textbf{x},t)$ of the RNN converges towards the true signal $\textbf{X}(\textbf{x},t)$ and thus the output $\textbf{y}_N$ can only become more accurate. This implies that provided there are two vertices $u, v$ such that $\textbf{Y}_u(T) \neq \textbf{Y}_v(T)$ and using Theorem \ref{thm:Universality},  $\mu(\textbf{y}_N)$ cannot converge to 0.


\textbf{Over-squashing} To formalize the concept of over-squashing, \cite{topping2021understanding} described it in terms of the impact of a distant node's feature \( \mathbf{x}_u \) on the prediction \( \mathbf{y}_v \) at node \( v \). They noted an exponential decay in \( \left|\frac{\partial \mathbf{y}_v}{\partial \mathbf{x}_u}\right| \) with increasing distance between nodes \( u \) and \( v \) in standard GNNs. Utilizing Theorem \ref{thm:Universality}, we find that the output $\mathbf{y}_v$ reacts similar to $\mathbf{Y}_v$ to changes along \( \mathcal{R}_v \) while being insensitive to perturbations perpendicular to \( \mathcal{R}_v \). This nuanced understanding enables a more targeted approach to addressing the limitations of GLAudio in handling long-range interactions.

\section{Experiments}
For details on training methods and hyper-parameters, we refer to Appendix \ref{app:ExperimentDetails}. The code for all experiments is available on \href{https://github.com/AurelioSulser/GLAudio}{https://github.com/AurelioSulser/GLAudio}.

\subsection{Node Classification on Network Graphs}
For a comparison with other model architectures, we evaluated the performance of GLAudio on six widely popular semi-supervised node classification single-graph datasets: Cora, CiteSeer, PubMed, Texas, Wisconsin and Cornell. Derived from citation networks, the first three are homophilic, i.e., adjacent vertices tend to have the same class label. The latter three are significantly more heterophilic, i.e. adjacent vertices tend to have different labels. Due to their smoothing bias, this property poses a significant challenge for traditional MPNNs like GCN. Average test accuracies for all six datasets are reported in Table \ref{tab:heterophilic_homophilic}. 

\textbf{Discussion and Results.}  We observe that on all six datasets GLAudio is able to successfully learn a classification strategy. On the heterophilic graph datasets, GLAudio outperforms the GCN and GAT architecture, whereas on homophilic graphs of Cora and CiteSeer GCN and GAT achieve higher accuracies. 
These results match our theoretic understanding: Both GCN and GAT can be seen as time discretizations of the discrete first-order heat equation equipped with learnable parameters causing them to naturally smooth the nodes' features. This bias is useful on homophilic datasets, yet disadvantageous on heterophilic graphs. Derived from the second-order wave equation, GLAudio does not exhibit this flaw, achieving consistently high accuracies on all datasets.

\begin{figure}[h]
    \vskip 0.1in
    \centering
    \includegraphics[width=\columnwidth]{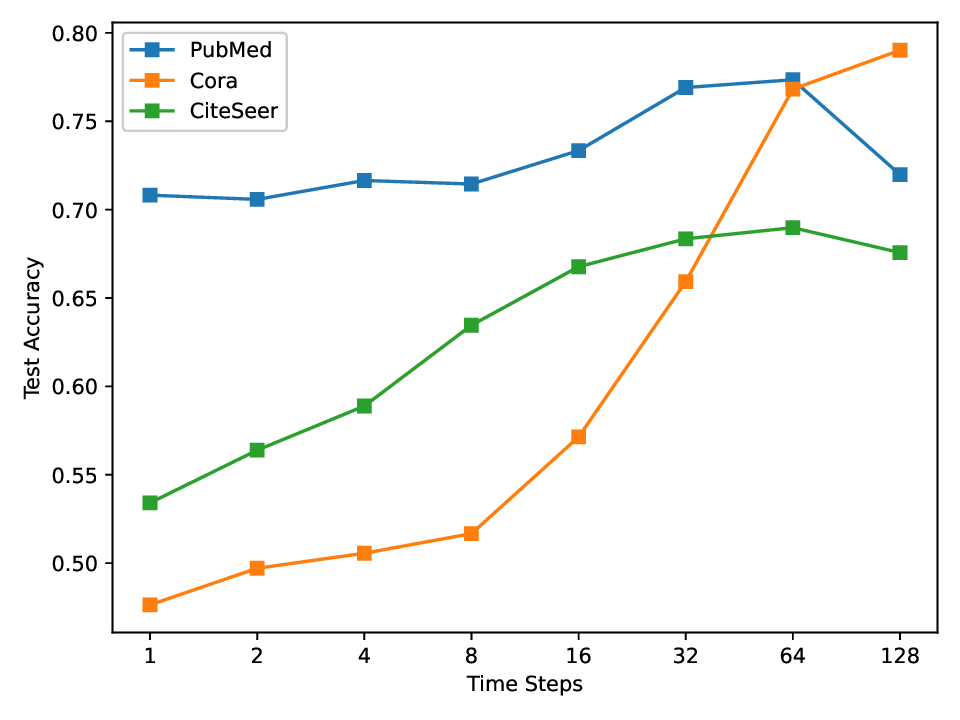}
    \caption{Test accuracy for varying numbers of time steps $N$. Accuracies are averaged over 10 random initializations.}
    \label{fig:impact-N-homophilic}
    \vskip -0.1in
\end{figure}

To verify that GLAudio is not prone to over-smoothing, we measured its accuracy on homophilic datasets by varying the time steps $N$, keeping other hyper-parameters constant, notably the stopping time $T$. Contrary to the effect of over-smoothing, GLAudio's performance improved with more time steps and reached optimal levels within the range of 50 to 200, as shown in Figure \ref{fig:impact-N-homophilic}. This contrasts with the findings in \cite{chamberlain2021grand} where GCNs' accuracies diminish significantly below 50\% with more than 16 layers.

\subsection{Graph Regression on ZINC Dataset}
In \cite{rusch2022graphcoupled}, it was found that the Mean Absolute Error (MAE) of standard GCNs on the ZINC dataset increased with the model's depth, while for GraphCon-GCN, the MAE showed a decreasing trend. This indicates the significance of long-range dependencies within the ZINC dataset. Given the capability of GLAudio to support exceptionally deep models, as highlighted in Figure \ref{fig:impact-N-homophilic}, its potential is explored on ZINC.

The ZINC dataset comprises approximately 12,000 molecular graphs, curated for the regression of a molecular property known as constrained solubility. Each graph represents a molecule, where the node features correspond to the types of heavy atoms present, and the edge features denote the types of bonds connecting these atoms.  

\begin{table}[h]
\caption{Test MAE on ZINC (without edge features, small $12 \mathbf{k}$ version) restricted to small network sizes of $\sim 100 k$ parameters. Baseline results are taken from \cite{beaini2021directional} and \cite{rusch2022graphcoupled}.}
\label{tab:Zinc}
\vskip 0.1in
\begin{center}
\begin{small}
\begin{sc}
\begin{tabular}{lc}
\hline 
Model & Test MAE \\
\hline 
GIN \cite{xu2018powerful} & $0.41$ \\
GatedGCN \cite{bresson2017residual} & $0.42$ \\
DGN \cite{beaini2021directional} & $0.22$ \\
\hline GCN \cite{kipf2017semisupervised} & $0.47$ \\
GraphCON-GCN \cite{rusch2022graphcoupled} & $0.22 $ \\
\hline GAT \cite{velickovic2017graph} & $0.46 $ \\
GraphCON-GAT \cite{rusch2022graphcoupled} & $0.23 $ \\
\hline
GLAudio-LSTM & $0.4342$ \\
\hline
\end{tabular}
\end{sc}
\end{small}
\end{center}
\vskip -0.1in
\end{table}

\textbf{Discussion and Results.} 
GLAudio's initial performance on the ZINC dataset appears modest, performing on par with GIN, and GatedGCN, see Table \ref{tab:Zinc}. The reason could be that GLAudio might disproportionately emphasize long-range over short-range interactions as models successful on ZINC typically emphasize the latter \cite{dwivedi2022benchmarking}. To test this, we combined a 4-layer GCN ($\sim$ 10,000 parameters) and a GLAudio-CoRNN ($\sim$ 90,000 parameters), allowing GCN to focus on short-range and GLAudio on long-range dependencies. The result achieved a test MAE of 0.3157, markedly surpassing standard GNN models and nearing state-of-the-art performance, suggesting GLAudio's proficiency in capturing dependencies missed by shallow GCN. 

\subsection{Long Range Graph Benchmark: Peptides-struct}
Long Range Graph Benchmark is a set of 5 datasets to measure the ability of GNNs and Graph Transformers to solve problems depending on long-range interactions \cite{dwivedi2023longrangegraphbenchmark}. We evaluated GLAudio on one of these tasks, namely Peptides-struct, a multi-dimensional graph regression task on molecular graphs.

\begin{table}[h]
    \caption{Test mean absolute error (MAE) on Peptides-struct. Baseline results are taken from \cite{dwivedi2023longrangegraphbenchmark}.}
    \label{tab:peptides-struct}
    \vskip 0.1in
\begin{center}
\begin{small}
\begin{sc}   
\begin{tabular}{lc}
\hline Model & Test MAE \\
\hline GCN \cite{kipf2017semisupervised} & $ 0.3496$ \\
GIN \cite{xu2018powerful} & $0.3547$ \\
GatedGCN \cite{bresson2017residual} & $0.3420$ \\
\hline
Transformer+LapPE \\ \cite{hamilton2017inductive} & $0.2529$ \\
\hline
GLAudio-LSTM & $0.3569$ \\
\hline
\end{tabular}
\end{sc}
\end{small}
\end{center}
\vskip -0.1in
\end{table}

\textbf{Discussion and Results.} The results in Table \ref{tab:peptides-struct} show GLAudio achieving on par MAEs with traditional MPNNs but underperforming graph transformers. This suggests that GLAudio might have difficulties capturing the long-range dependencies of Peptides-struct. On the contrary, we observed significant performance enhancements with an increasing number of time steps. Moreover, recent findings from \cite{tönshoff2023did} reveal that a simple GCN with extensive hyper-parameter tuning can achieve state-of-the-art results (0.2460 MAE) on Peptides-struct. It casts doubt on the suitability of Peptides-struct as a benchmark for long-range graph interactions.

\section{Conclusion}
We proposed a novel graph learning architecture based on wave propagation. A central distinction from other models lies in the separation of information propagation and information processing into two different steps. This separation allows for deep feature propagation without over-smoothing.
Moreover, our theoretical study of expressivity provides a new approach to understanding over-squashing as a miss-alignment between graph structure and task. In our empirical studies, GLAudio was benchmarked against state-of-the-art models using network datasets and molecular graphs. On heterophilic datasets, we empirically validate that GLAudio alleviates over-smoothing. Additionally, we investigated the phenomenon of over-squashing on more complex datasets, namely ZINC and Peptides-struct. Based on our results, it remains inconclusive as to whether GLAudio significantly mitigates over-squashing. Despite not yet achieving state-of-the-art scores on more complex datasets, we believe that our approach of separating information processing from feature propagation represents a promising concept that leaves space for further investigation.

\nocite{langley00}

\bibliography{example_paper}
\bibliographystyle{icml2024}

\newpage
\appendix
\onecolumn

\section{Proof of Theorem \ref{thm:GraphEncoding}}
We remark that the power series representation of the signal $\mathbf{X}_H(t)$ is given by
    \[\mathbf{X}_H(t) = \cos(\mathbf{L}_H^{1/2} \cdot t) \cdot \mathbf{x}_H = \sum_{n=0}^{\infty} \frac{(-1)^n \mathbf{L}_H^n t^{2 n}}{(2 n) !} \cdot \mathbf{x}_H = \sum_{n=0}^{\infty} \frac{(-1)^n \cdot (\mathbf{L}_H^n \cdot \mathbf{x}_H) \cdot t^{2 n}}{(2 n) !}.\]
It is thus immediate that $\forall n \in \N_0: \mathbf{L}_H^n \cdot \mathbf{x}_{H} = \mathbf{L}_G^n \cdot \mathbf{x}_{G}$ implies that
\[\mathbf{X}_H(t) = \mathbf{X}_G(t).\]
For the other implication, we note that $ \mathbf{X}_{G}|_{[0,t]} = \mathbf{X}_{H}|_{[0,t]}$ implies that 
\[\forall n \in \N_0: \frac{d^{(2n)}}{dx^{(2n)}} \mathbf{X}_{G}(0) = \frac{d^{(2n)}}{dx^{(2n)}}\mathbf{X}_{H}(0) .\]
According to the power series representation of $\mathbf{X}_G(t)$ and $\mathbf{X}_H(t)$, this is equivalent to
\[\mathbf{L}_H^n \cdot \mathbf{x}_{H} = \mathbf{L}_G^n \cdot \mathbf{x}_{G}.\]

\section{Proof of Theorem \cref{thm:Universality}}\label{app:Expressivity}

Before we present the actual proof, we introduce the following definition.

We call an operator $\Phi:\left(L^{\infty}\left([0, 1] ; \mathbb{R}\right),\|\cdot\|_{L^{\infty}}\right) \rightarrow\left(L^{\infty}\left([0, 1] ; \mathbb{R}\right),\|\cdot\|_{L^{\infty}}\right)$ causal if for any two input signals $u, v \in L^{\infty}\left([0, 1] ; \mathbb{R}\right)$ 
\[u|_{[0, t]} = v|_{[0, t]} \longrightarrow \Phi(u)|_{[0, t]}=\Phi(v)|_{[0, t]}.\]

We begin the proof with the claim that if there exists a continuous (with respect to the $L^{\infty}$-norm on the input-/output-signals), causal operator $\Phi$ such that $\Phi(\textbf{X}_v)(T) = \textbf{Y}_v(\textbf{x})$, then we can conclude.

\begin{proof}[Proof of the claim.]
    As a preliminary step, we define $K=\{\textbf{X}^i_v(\textbf{x}) | \forall 1 \leq i \leq N, \text{ for all initial configuration }\mathbf{x}\in C\}$. It follows from compactness of $C$ that $K$ is compact. In the following, we will give a description of $\Phi(\hat{\textbf{X}}_v)(T)$ as a dynamical system and apply Theorem \ref{thm:RNN-universality} to this description together with the compact set $K$. 
    
    For an arbitrary input sequence $\textbf{X}_1, \dots, \textbf{X}_N$, set $\textbf{S}_i = (i, \textbf{X}_i, \textbf{V}_i)$, where $\textbf{V}_i = \textbf{V}_{i-1} - h \cdot \textbf{L} \cdot \textbf{X}_{i-1}$ and $\textbf{V}_0 = 0$. It is clear that $\textbf{S}_{i+1}$ is computable from $\textbf{S}_i, \textbf{X}_{i+1}$, hence there exists a measurable function $g$ such that $\textbf{S}_{i+1} = g(\textbf{S}_i, \textbf{X}_{i+1})$. We note that for the special input sequence $\textbf{X}_1 = \textbf{X}^1_v, \dots, \textbf{X}_N = \textbf{X}^N_v$, we have that $\textbf{S}_i = (i, \textbf{X}^i_v, \textbf{V}^i_v)$.
    
    To define the function $h(\textbf{S}_i)$, let us introduce vectors $\tilde{\textbf{X}}_1, \dots, \tilde{\textbf{X}}_i, \tilde{\textbf{V}}_1, \dots, \tilde{\textbf{V}}_i$ (that are in some sense approximations of $\textbf{X}_1, \dots, \textbf{X}_i, \textbf{V}_1, \dots, \textbf{V}_i$) given by 
    \begin{align*}
    \tilde{\textbf{X}}_{j} &= \tilde{\textbf{X}}_{j+1} - h \tilde{\textbf{V}}_{j+1}\\
    \tilde{\textbf{V}}_{j} &= \tilde{\textbf{V}}_{j+1} + h \cdot \textbf{L}\cdot \tilde{\textbf{X}}_{j}\\
    \tilde{\textbf{X}}_i &= \textbf{X}_i \quad \tilde{\textbf{V}}_i = \textbf{V}_i.
    \end{align*}
    In the special case where $\textbf{X}_1 = \textbf{X}^1_v, \dots, \textbf{X}_N = \textbf{X}^N_v$, we have that $\tilde{\textbf{X}}_1 = \textbf{X}^1_v, \dots, \tilde{\textbf{X}}_N = \textbf{X}^N_v$.
    We define the continuous function $f: \R^{|V(G)|} \times \R^{|V(G)|} \rightarrow L^{\infty}([0,1];\R)$ given by $f(\textbf{X}_i, \textbf{V}_i)(t) = \tilde{\textbf{X}}_i \cdot 1_{[0,h]}(t) + \sum_{j = 2}^i \tilde{\textbf{X}}_i \cdot 1_{((i-1)h,ih]}(t)$. This allows us to define $h(\textbf{S}_i) = \Phi(f(\textbf{X}_i, \textbf{V}_i))(ih).$ We note that in the special case that $\textbf{X}_1 = \textbf{X}^1_v, \dots, \textbf{X}_N = \textbf{X}^N_v$ that $f(\textbf{X}_i, \textbf{V}_i)|_{[0,ih]} = \hat{\textbf{X}}_v|_{[0,ih]}$ and hence by causality of $\Phi$, we have that $h(\textbf{S}_i) = \Phi(\hat{\textbf{X}}_v)(ih).$ Applying Theorem \ref{thm:RNN-universality} to this dynamic system with compact set $K$, we get the output of some RNN $\textbf{y}_N$ satisfies that $\forall \textbf{x} \in C: \|\Phi(\hat{\textbf{X}}_v(\textbf{x}))(Nh) - \textbf{y}_N(\textbf{X}^1_v(\textbf{x}), \dots, \textbf{X}^N_v(\textbf{x}))\| \leq \epsilon$. Provided we picked $N$ large enough such that $\forall \textbf{x} \in C: \|\textbf{Y}_v(\textbf{x}) - \Phi(\hat{\textbf{X}}_v(\textbf{x}))(Nh)\| = \|\Phi(\textbf{X}_v(\textbf{x}))(Nh) - \Phi(\hat{\textbf{X}}_v(\textbf{x}))(Nh)\| \leq \epsilon$, we can combine the two bounds to obtain the statement of the theorem.
    
\end{proof}

All that remains is to argue the existence of such a continuous, causal operator $\Phi$. Let us make the following preliminary remarks. Let $\phi_1, \dots, \phi_n$ be normalized eigenvectors of $\mathbf{L}$ associated with eigenvalues $\lambda_1, \dots, \lambda_n$. W.l.o.g. we may assume that $\lambda_1, \dots, \lambda_n \in \Q$ and let $k \in \N_1$ such that $\forall 0 \leq i \leq n: k \cdot \lambda_i \in \N$. Moreover, we denote the spectral decomposition of $\mathbf{L}$ by $\mathbf{U} \cdot \mathbf{\Lambda} \cdot \mathbf{U}^t,$ where $\mathbf{U}$ is orthonormal and $\mathbf{\Lambda}$ is diagonal, and for all $i \geq 1$ we introduce the functions
\[u_i(t) = \frac{\cos(\lambda_i \cdot t)}{\langle \mathbf{e}_v, \phi_i \rangle}.\]
We write $\langle \cdot , \cdot \rangle_t$, where $t > 0$, for the scalar product defined on $C_0([0,1], \R)$ by
\[\langle u , v \rangle_t = \frac{1}{\pi k} \int_{[0, 2\pi \cdot k \cdot t]} u\left(\frac{s}{2\pi \cdot k}\right) \cdot v\left(\frac{s}{2\pi \cdot k}\right) ds \]
We remark that given the wave signal $\mathbf{X}(t) = \sum_{j=1}^n \phi_j \cos(\lambda_j \cdot t) \langle \phi_j , \mathbf{x} \rangle$ corresponding to the initial configuration $\mathbf{x}$, the scalar product satisfies 
\begin{align}
    \langle u_i, \textbf{X}_v \rangle_1 &= \sum_{\phi_j \in \mathcal{R}_v} \langle u_i, \langle \phi_j, \mathbf{e}_v \rangle \cdot \cos(\lambda_j \cdot t) \cdot \langle \phi_j, \mathbf{x} \rangle \rangle_1 \\
    &= \sum_{\phi_j \in \mathcal{R}_v} \langle \phi_j, \mathbf{x} \rangle \cdot \frac{\langle \phi_j, \mathbf{e}_v \rangle}{\langle \phi_i, \mathbf{e}_v \rangle} \cdot\langle \cos(\lambda_i \cdot t), \cos(\lambda_j \cdot t) \rangle_1 \\
    &= \sum_{\phi_j \in \mathcal{R}_v} \langle \phi_j, \mathbf{x} \rangle \cdot \delta_{ij}.
\end{align}
With these preparations at hand, it is easy to conclude. We note that since $\textbf{Y}_v$ is supported on $\mathcal{R}_v$, the function $\textbf{Y}_v$ factorizes into $\textbf{Y}_v = \tilde{\textbf{Y}}_v \circ p $, where $p: \R^n \rightarrow \R^{|\mathcal{R}_v|}$ is the projection
\[p(\mathbf{x}) = \left(\langle \mathbf{x}, \phi_i \rangle\right)_{i \geq 1: \phi_i \in \mathcal{R}_v} \]
and $\tilde{\textbf{Y}}_v: \R^{|\mathcal{R}_v|} \rightarrow \R$ is continuous.

We define the operator $\Phi: C_0\left([0, T] ; \mathbb{R} \right) \rightarrow C_0\left([0, T] ; \mathbb{R} \right)$ by 
\[ \Phi(u)(t) = \tilde{\textbf{Y}}_v((\langle u,  u_i \rangle_t)_{i \geq 1: \phi_i \in \mathcal{R}_v}).\]
We remark that $\Phi$ is causal and continuous. Moreover, given the wave signal $\mathbf{X}(t)$ corresponding to the initial configuration $\mathbf{x}$, we have that
\[\Phi(\textbf{X}_v)(T) = \tilde{\textbf{Y}}_v((\langle \textbf{X}_v, u_i \rangle_1)_{i\geq 1: \phi_i \in \mathcal{R}_v}) = \tilde{\textbf{Y}}_v((\langle \mathbf{x}, \phi_i \rangle)_{i\geq 1: \phi_i \in \mathcal{R}_v}) = \tilde{\textbf{Y}}_v \circ p(\mathbf{x}) = \textbf{Y}_v(\mathbf{x}).\] 

\section{Further Experiment and Training Details 
\label{app:ExperimentDetails}}

Our base architecture admits the following hyper-parameters: Number of layers $L$, number of time steps $N$, step size $h$, hidden dimension size $q$, stopping time $T$, and the choice of activation function $\sigma$. Note that $N, h$, and $T$ are coupled by $T = Nh$.

Further, we experimented with initial node-wise linear or non-linear embeddings, different dropout rates, and varying constant or adaptive (ReduceOnPlateau) learning rates. The eigenvalues of the Laplacian $\textbf{L}$, i.e., the frequencies of our oscillation $\textbf{X}(t)$ lie in $[0, 2 \max_v\mathrm{deg}(v)]$. Let $\textbf{D}$ denote the degree matrix. To avoid very high frequencies, difficult to be captured by our numerical approximation of $\textbf{X}(t)$, we also considered normalizing the Laplacian in \eqref{eq:Wave} by $\textbf{N} = \textbf{D}^{-\frac{1}{2}}\textbf{L}\textbf{D}^{-\frac{1}{2}}$. The eigenvalues of $\mathbf{N}$ lie in $[0,2]$. Similarly to avoid very slow frequencies, we implemented the option of adding self-loops, leading to $\textbf{I} + \textbf{L}$ or $\textbf{I} + \textbf{N}$ instead of $\textbf{L}$ resp. $\textbf{N}$ in \eqref{eq:Wave}.

We used the Adam optimizer with various learning rates and weight decays. 
For node classification tasks, we used the multi-class cross-entropy function as a loss function. For regression tasks, we used the $\ell1$-function as the loss function.

In our training process, each training run of GLAudio for all six network datasets was conducted over a span of 300 epochs. For ZINC the maximum number of epochs was 1000 and for Peptides-struct 300. For ZINC and Peptides-Struct early-stopping at stagnating validation loss was used to avoid overfitting.

\begin{table*}[t]
\caption{Overview of best-found hyperparameter configurations for all experiments.}
\label{tab:hyperparameters}
\vskip 0.1in
\centering 
\small
\begin{tabular}{lcccccccccc}
\toprule
Dataset & L  & N  & h &  \begin{tabular}[c]{@{}c@{}}Learning\\ Rate\end{tabular} & \begin{tabular}[c]{@{}c@{}}Weight \\Decay\end{tabular}  & Activation & \begin{tabular}[c]{@{}c@{}} Hidden \\Dim. \end{tabular} & \begin{tabular}[c]{@{}c@{}} Normalized \\Laplacian \end{tabular}  & \begin{tabular}[c]{@{}c@{}}Dropout \\ Rate\end{tabular} & \begin{tabular}[c]{@{}c@{}}Self \\ Loops\end{tabular} \\
\midrule
Cora & 2 & 200 & 0.02 & 0.001 & 0.005 & Leaky ReLU & 32 & True &0.2&False\\
Citeseer  & 1 & 150 & 0.01 & 0.0025 & 0.005 & Leaky ReLU & 24 & True & 0.2 &False\\
Pubmed  & 1 & 50 & 0.06 & 0.0025 & 0.005 & Leaky ReLU & 24 & False &0.3&False\\
Texas & 2 & 10 & 0.15 & 0.001 & 0.01 & Leaky ReLU & 32 & False & 0.4  &False\\
Cornell & 1 & 2 & 0.7 & 0.0025 & 0.005 & ReLU & 48 & True & 0.4&False\\
Wisconsin & 1 & 4 & 0.2 & 0.01 & 0.001 & Leaky ReLU & 32 & True &0.3 &False\\
ZINC (CoRNN + GCN)&  3 & 100 & 0.02 & 0.005 & 0.0 & Leaky ReLU & 150 + 40 & True & 0.0 & True \\
ZINC (LSTM) & 3 & 20 & 0.5 & 0.005 & 0.0 & ReLU & 128 & True & 0.0 & True\\
Peptides-struct (LSTM) & 3 & 20 & 0.5 & 0.002 & 0.0 & GeLU & 128 & True & 0.0 & False\\
\bottomrule
\end{tabular}

\end{table*}


\end{document}